\documentclass[runningheads]{llncs}

\PassOptionsToPackage{table,xcdraw}{xcolor}
\usepackage{eccv}

\usepackage{eccvabbrv}
\usepackage{wrapfig}
\usepackage{graphicx}
\usepackage{booktabs}
\usepackage{multirow}
\usepackage[accsupp]{axessibility}  
\usepackage[ruled,vlined]{algorithm2e}
\usepackage{amsmath}
\usepackage{tikz}
\usepackage{caption}

\graphicspath{{figs/}}
\newcommand{\mysection}[1]{\vspace{2pt}\noindent\textbf{#1}}

\definecolor{myblue}{HTML}{478feb}
\definecolor{myred}{HTML}{fb4337}
\definecolor{mygreen}{HTML}{6cc794}
\definecolor{mypurple}{HTML}{9c45ef}
\definecolor{mygray}{HTML}{302c2c}

\newcommand{\coloredcircled}[2]{%
  \tikz[baseline=(char.base)]{
    \node[shape=circle, fill=#1, inner sep=.5pt] (char) {\textcolor{white}{#2}};}}

\SetAlCapSty{textbf}
\SetAlCapNameFnt{\scriptsize}
\SetAlCapFnt{\scriptsize}
\definecolor{commentcolor}{RGB}{140,140,140}
\definecolor{rnamecolor}{RGB}{238,129,195}
\definecolor{fnamecolor}{RGB}{0,155,0}

\newcommand{\PyComment}[1]{\fontfamily{pcr}\selectfont\textcolor{commentcolor}{\# #1}}

\newcommand{\PyCode}[1]{\fontfamily{pcr}\selectfont #1} 

\newcommand{\colormae}{\gradientRGB{Color}{71,143,235}{73,187,121}\gradientRGB{MAE}{156,69,239}{251,67,55}}

\newcommand{\greencolormae}{\textcolor{mygreen}{ColorMAE-G~}}

\usepackage{hyperref}

\usepackage{orcidlink}

\usepackage{gradient-text}

\begin{document}
\title{\texorpdfstring{\gradientRGB{Color}{71,143,235}{73,187,121}\gradientRGB{MAE}{156,69,239}{251,67,55}}{ColorMAE}: Exploring data-independent masking strategies in Masked AutoEncoders}

\titlerunning{ColorMAE}
\author{Carlos Hinojosa \orcidlink{0000-0001-9286-9587} \and
Shuming Liu \orcidlink{0000-0001-5227-647X} \and
Bernard Ghanem \orcidlink{0000-0002-5534-587X}}

\authorrunning{C.~Hinojosa et al.}

\institute{King Abdullah University of Science and Technology (KAUST) \\
\url{https://carloshinojosa.me/project/colormae}}


\maketitle

\begin{abstract}

Masked AutoEncoders (MAE) have emerged as a robust self-supervised framework, offering remarkable performance across a wide range of downstream tasks. 
To increase the difficulty of the pretext task and learn richer visual representations, existing works have focused on replacing standard random masking with more sophisticated strategies, such as adversarial-guided and teacher-guided masking. However, these strategies depend on the input data thus commonly increasing the model complexity and requiring additional calculations to generate the mask patterns.
This raises the question: \textit{Can we enhance MAE performance beyond random masking without relying on input data or incurring additional computational costs?} In this work, we introduce a simple yet effective data-independent method, termed \colormae, which generates different binary mask patterns by filtering random noise. Drawing inspiration from color noise in image processing, we explore four types of filters to yield mask patterns with different spatial and semantic priors.
\colormae \ requires no additional learnable parameters or computational overhead in the network, yet it significantly enhances the learned representations. We provide a comprehensive empirical evaluation, demonstrating our strategy's superiority in downstream tasks compared to random masking. Notably, we report an improvement of 2.72 in mIoU in semantic segmentation tasks relative to baseline MAE implementations.
\end{abstract}


\section{Introduction}
\label{sec:intro}

Self-supervised learning (SSL) has emerged as a prominent pre-training paradigm, favored for its capacity to learn rich representations without the need for human-labeled data \cite{oord2018representation,ermolov2021whitening}. Recent advancements demonstrate that large-scale SSL significantly outperforms supervised learning on challenging datasets. Inspired by masked language modeling (MLM) \cite{brown2020language,devlin2018bert} in natural language processing and the development of vision transformers (ViT) \cite{dosovitskiy2020image}, masked image modeling (MIM) has achieved outstanding downstream performance across a broad spectrum of computer vision tasks \cite{bao2021beit,he2022masked}, thereby attracting increasing attention.

\begin{figure}[t]
  \centering
  \includegraphics[width=\columnwidth]{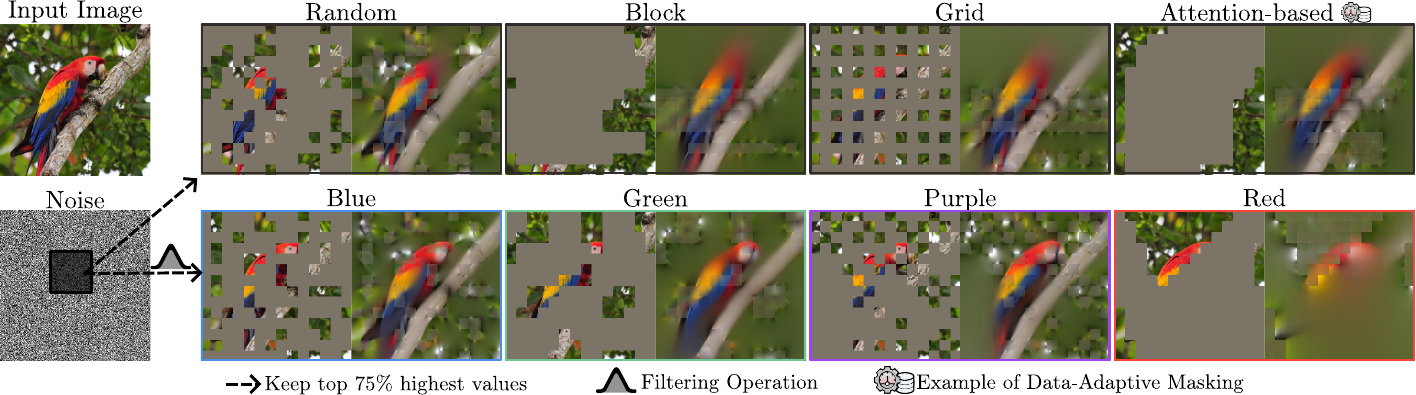}
  \caption{\small We use the MAE\cite{he2022masked} to mask and reconstruct an input image using different masking strategies with a masking ratio of 75\%. The results of the first three columns shown in the top row correspond to traditional data-independent strategies: random, block-wise, and grid-wise masking, respectively. The last column of the top row shows an example of adaptive masking using an attention-based mechanism. The second row shows our four distinct types of masks generated by \colormae~when filtering random noise with high-pass (Blue), band-pass (Green), band-stop (Purple), and low-pass (Red) filters.}
  \label{fig:intro}
\end{figure}

MIM learns rich representations during pre-training by masking certain patches of the input image and predicting their original content based on the remaining unmasked patches. In this context, the strategy of masking plays a pivotal role. Current strategies fall into two categories: \textit{data-independent masking} and \textit{data-adaptive masking}. 
The former includes conventional random masking~\cite{he2022masked}, block-wise masking\cite{bao2021beit}, which masks contiguous blocks of image patches, and grid-wise masking, which keeps one out of every four patches, as shown in the first row of Fig. \ref{fig:intro}. To date, random masking has proven effective for most MIM methods due to its higher masking ratio and the implementation simplicity~\cite{he2022masked}. 
The latter category, \textit{adaptive masking strategies}, involves designing mask patterns based on additional information, feedback, or image context analysis \cite{kakogeorgiou2022hide,li2022semmae}, as illustrated in the first row, last column of Fig. \ref{fig:intro}. While adaptive masking often excels in downstream tasks, it necessitates the incorporation of attention-driven mechanisms or additional feedback, thereby increasing computational costs. Conversely, data-independent masking strategies, though simpler and without extra computational demands, have been scarcely explored beyond random masking.

In this work, we advocate for data-independent masking and propose a novel, straightforward, yet effective noise-filtering masking strategy. This strategy maintains the simplicity advantage of conventional random masking while facilitating the learning of stronger visual representations. 
Specifically, MAE initially generates a random noise array and selects the patches according to the desired mask ratio. From the signal/image processing perspective, the random noise array could be generated from white noise \cite{castleman1996digital}, characterized by a flat frequency spectrum with equal power across all bands. Considering noise frequency analysis and inspired by the concept of color noise in image processing \cite{lau2003blue,correa2016spatiotemporal}, we propose to generate mask patterns with varied constrained spectra, named \colormae. By filtering random noise through low-pass, high-pass, band-pass, and band-stop filters, we produce different noise patterns labeled as \textcolor{myred}{red}, \textcolor{myblue}{blue}, \textcolor{mygreen}{green}, and \textcolor{mypurple}{purple}, respectively. These variants possess distinct properties and frequency spectra, and show different spatial and semantic priors, as depicted in the second row of Fig. \ref{fig:intro}.

Our method, which does not rely on external guidance or additional learnable parameters, maintains computational efficiency during pre-training similar to random masking. Our extensive experiments reveal that one of our innovative masking strategies enables MAE to learn superior image representations during pre-training, surpassing conventional random masking in three downstream vision tasks: image classification, semantic segmentation, and object detection. Notably, our approach achieves a significant increase in mean Intersection over Union (mIoU) by 2.72 in semantic segmentation compared to random masking. We hope our approach could be used to design new MIM architectures or as a foundational masking strategy for developing adaptive masking techniques.

\vspace{4pt}
\mysection{Contributions}. We summarize our contributions as follows:
\begin{itemize}
    \item[(i)] We propose a simple yet effective masking strategy to generate different data-independent masks by sampling and filtering random noise. Our method does not incorporate additional learnable parameters into the MAE model, preserving computational efficiency during pre-training.
    \item[(ii)] We investigate four distinct mask types created by applying low-pass, high-pass, band-pass, and band-stop filters to random noise. We offer detailed analysis and comparisons of these masks across three downstream tasks: image classification, semantic segmentation, and object detection.
    \item[(iii)] Through extensive experiments, we demonstrate that the ``Green masking'' (\textcolor{mygreen}{ColorMAE-G}), achieved by applying a band-pass filter to random noise, significantly enhances MAE performance compared to random masking.
\end{itemize}





\section{Related Works}
\label{sec:related_works}

\subsection{Self-supervised Learning}

\mysection{Contrastive Learning.} Among the numerous self-supervised learning approaches in computer vision that focus on learning from unlabeled data, contrastive learning has emerged in recent years \cite{wu2018unsupervised, oord2018representation, hjelm2018learning, bachman2019learning, grill2020bootstrap, chen2021exploring}. Its fundamental principle involves learning representations through instance discrimination, which involves attracting similar samples while optionally repelling dissimilar ones. SimCLR \cite{chen2020simple}, a prominent method in this domain, enhances representations by maximizing the similarity between different views of the same image with large training batches. MoCo \cite{he2020momentum} advances this approach by employing a memory bank and a momentum-updated encoder to complement the pool of negative samples, thereby learning more robust representations. DINO \cite{caron2021emerging} incorporates a self-distillation mechanism, compelling the student network to mimic the teacher network's output on augmented views of the same image, which fosters strong attention to the salient parts of images.

\mysection{Masked Image Modeling.} Inspired by the success of Masked Language Modeling (MLM) \cite{devlin2018bert} in natural language processing, Masked Image Modeling (MIM) has garnered increasing interest in the vision domain \cite{chen2024context, he2022masked, zhou2021ibot, li2021mst, baevski2022data2vec, el2021large}. This approach aims to reconstruct the original image from masked inputs, with reconstruction targets varying from raw pixels \cite{he2022masked, xie2022simmim, chen2023mixed} and dVAE tokens \cite{bao2021beit} to HoG features \cite{wei2022masked} and frequency components \cite{xie2022masked}. Notably, MAE \cite{he2022masked} has achieved significant attention for its simplicity and computational efficiency. It introduces an asymmetric encoder-decoder architecture, where the encoder processes only a subset of visible patches selected through a random masking strategy, while a lightweight decoder predicts the original image using both masked and unmasked patches. Pretraining with MAE has been shown to substantially enhance performance on downstream tasks compared to supervised pretraining. This work builds upon MAE's framework and proposes novel masking strategies that improve the random masking approach.

\subsection{Masking Strategy}

\mysection{Data-Adaptive Masking.} The selection of mask sampling strategies is pivotal in MIM as it defines the difficulty of the pretext task, thereby influencing both the quality of the reconstruction and the learned representations. Recent studies have explored more advanced masking strategies. For example, AttMask \cite{kakogeorgiou2022hide} selects patches for masking based on high scores in the teacher network's attention map, presenting a more challenging pretext task. ADIOS \cite{shi2022adversarial} utilizes adversarial training to increase the difficulty of the pretext task. SemMAE \cite{li2022semmae} targets semantic portions of the image, masking patches within these areas to provide a nuanced challenge. HPM \cite{wang2023hard} posits that the difficulty of MIM reconstruction can be quantified by patch-wise reconstruction loss, leading to the development of an auxiliary loss predictor for strategic masking. Feng \cite{feng2023evolved} introduces an evolved masking strategy that incrementally focuses on object semantics and context by effectively masking precise object parts. These strategies, which depend on the image's pixel values for mask sampling, are collectively referred to as \textit{data-adaptive masking}. This term signifies that the masking process is conditioned and adaptively chosen based on the input data.

\mysection{Data-Independent Masking.} Conversely, a category of masking techniques exists that does not depend on the input images or external guidance, termed \textit{data-independent masking}. Among these, random masking stands out for its simplicity and is implemented in MAE \cite{he2022masked} and SimMIM \cite{xie2022simmim} with a large masking ratio. Block-wise masking, adopted by BEiT \cite{bao2021beit} and BootMAE \cite{dong2022bootstrapped}, involves masking contiguous blocks of image patches. Furthermore, grid masking—a strategy that regularly obscures a grid pattern across the image—has been proposed as a data augmentation method in \cite{chen2020gridmask} and explored in MAE. Our work extends the data-independent masking framework by applying different filters on random noise, thus achieving computational efficiency with a fast-speed masking function. This proposed methodology not only enhances visual representation learning but also provides substantial benefits for downstream tasks, offering an improvement over traditional random masking approaches.





\begin{figure}[t]
    \centering
    \includegraphics[width=\columnwidth]{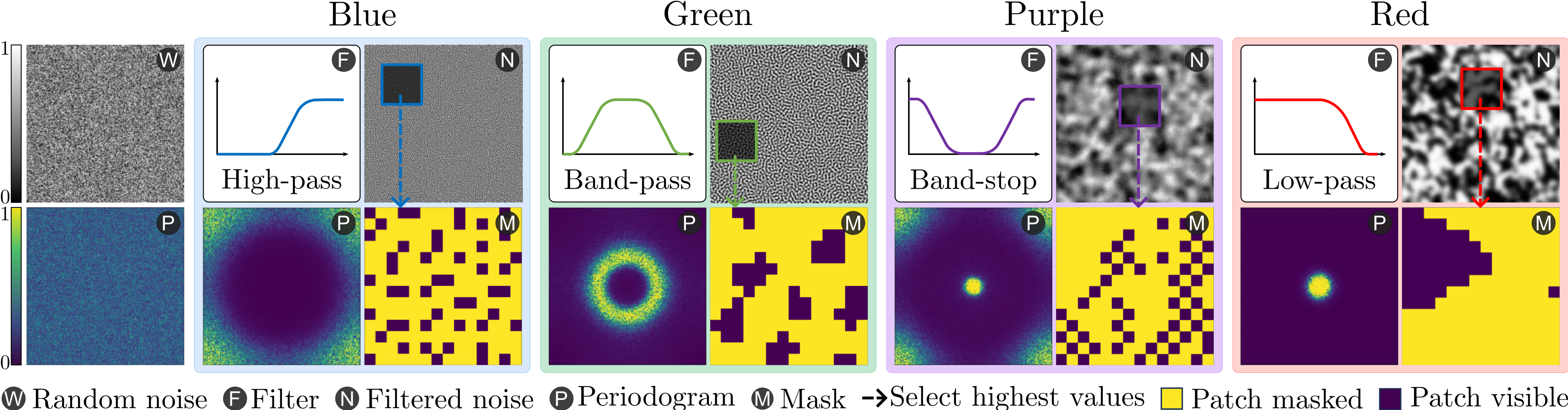}
    \caption{Starting with random noise \protect\coloredcircled{darkgray}{\scriptsize W}, we apply four filters \protect\coloredcircled{darkgray}{\scriptsize F}: high-pass (Blue), band-pass (Green), band-stop (Purple), and low-pass (Red) to produce the filtered noises \protect\coloredcircled{darkgray}{\scriptsize N}. The periodogram \protect\coloredcircled{darkgray}{\scriptsize P} of each filtered noise is displayed in the second row; as observed, each version of \protect\coloredcircled{darkgray}{\scriptsize N} exhibits a distinctive pattern in the frequency domain. Once the filtered noise is obtained, we perform a random crop on \protect\coloredcircled{darkgray}{\scriptsize N} to obtain a local window (sized to match the total number of patches) and select the top values according to the desired mask ratio (\eg, 75\%) to create the binary mask \protect\coloredcircled{darkgray}{\scriptsize M} used during pre-training.}
    \label{fig:proposed}
\end{figure}


\section{Proposed Method}
\label{sec:method}
To efficiently create binary masks during pre-training, MAE~\cite{he2022masked} generates random uniform noise and then selects the data points with the highest top values according to the desired mask ratio (e.g., 75\%). These selected values determine which patches are masked out (represented by ones in the binary mask) or remain visible in the input image, hence directing which portions of the input data the model will attempt to reconstruct. In this regard, the mask is critical in determining how high-level semantic representations are extracted from low-level features like image pixels.


In addition to uniform noise, \textit{white noise}~\cite{castleman1996digital} can also be implemented to produce random patterns. In general, uniform and white noise are not identical, as their mathematical definitions and statistical properties differ. For instance, white noise has an evenly distributed signal across the frequency spectrum, while uniform noise does not inherently have a flat spectral distribution. However, in practical experiments, they exhibit similar behavior and generate comparable random masks. Consequently, there are no significant differences in MAE pre-training and fine-tuning performance when incorporating either noise type; see our supplementary material for further details. Therefore, we will refer to both white and uniform noise simply as \textit{random noise} throughout the manuscript.



Moreover, beyond white noise, there are other types of noise in the image processing field, known as \textit{color noise}\cite{vasseur2004color, lau2003blue,correa2016spatiotemporal}. Unlike white noise, which maintains consistent power across its frequency bands, color noises exhibit unique spectral distributions, such as a predominance in the low-frequency band. Although color noise has been investigated in other domains, its application within deep learning frameworks, especially self-supervised learning, remains largely unexplored.

In this work, we draw inspiration from the concept of color noise in image processing and introduce a novel approach, termed \colormae, which employs mask patterns with distinct spectral constraints to facilitate efficient self-supervised learning. Concretely, instead of directly using random noise, our method applies different filters to random noise, \eg low-pass, high-pass, band-pass, and band-stop, to generate diverse noise patterns that embody unique spatial and frequency characteristics, as illustrated in Fig. \ref{fig:proposed}. This masking strategy is independent of data, eliminating the need for additional instructional inputs or loss functions, thus ensuring efficient and rapid sampling akin to MAE's random masking. However, our experiments demonstrate that certain color noise can significantly enhance visual representation quality during pre-training. To align with traditional terminology in image processing, we categorize the produced noise patterns as \textcolor{myred}{Red}, \textcolor{myblue}{Blue}, \textcolor{mygreen}
{Green}, and \textcolor{mypurple}{Purple} noise. Next, we will formally detail the definitions and implementations of these color noise masking.


\vspace{4pt}
\mysection{\textcolor{myred}{Red Noise.}} Let $W(x,y)$ represent a random noise image, where $x$ and $y$ are spatial coordinates. We apply a blurring operation over $W$ using a Gaussian kernel $G_{\sigma}$ with standard deviation $\sigma$ to filter out the high-frequency components and accentuate low frequencies effectively. This operation transforms the random noise into \textit{\textcolor{myred}{red noise}} $N_r$ given by:
\begin{equation}
    N_r=G_{\sigma}*W,
\end{equation}
\noindent where $*$ denotes the convolution operation. Finally, we perform normalization on $N_r$ to adjust the intensity values accordingly. We iteratively repeat the low-pass filtering and normalization steps to refine the noise characteristics.

\vspace{4pt}
\mysection{\textcolor{myblue}{Blue Noise.}} To generate blue noise patterns, it is required to apply a high-pass filter over $W$. A practical approach to implementing a high-pass filter involves first applying a low-pass filter ($G_{\sigma}*W$) to obtain the low-frequency content. Then, this filtered output is subtracted from the original random noise image $W$, effectively retaining the high-frequency components. The resulting \textit{\textcolor{myblue}{blue noise}} $N_b$ is formally expressed as
\begin{equation}
    N_b=W-G_{\sigma}*W.
\end{equation}
Note that alternative algorithms, such as the Void and Cluster method \cite{ulichney1993void}, can also be employed to generate high-quality blue noise patterns. This algorithm initiates with a random distribution of points and iteratively adjusts their placement to fill gaps evenly while avoiding the formation of clusters. It computes a density metric for empty spaces based on nearby points, placing new points strategically for even distribution. This algorithm and the blue noise patterns have been extensively used in the computer graphics field \cite{wolfe2022spatiotemporal, ahmed2020screen}.

\vspace{4pt}
\mysection{\textcolor{mygreen}{Green Noise.}} This noise is defined as the mid-frequency component of white noise; \ie, it can be generated by applying a band-pass filter over $W$ to eliminate both high and low frequencies. Such band-pass filtering effect can be approximated by sequentially applying two Gaussian blurs: first, a weak blur is applied to $W$ to remove the highest frequency details, followed by a separate strong blur to capture the lowest frequency content of $W$. By subtracting the strongly blurred version of $W$ from the weakly blurred one, the resultant noise image retains only the mid-frequency components. Formally, the \textit{\textcolor{mygreen}{green noise}} $N_g$ image can be obtained as:
\begin{equation}
N_g=G_{\sigma_1}*W -G_{\sigma_2}*W,
\end{equation}
\noindent where $\sigma_1$ and $\sigma_2$ denote the standard deviation of the two Gaussian kernels with  $\sigma_1 < \sigma_2$.

\vspace{4pt}
\mysection{\textcolor{mypurple}{Purple Noise.}} Finally, in this paper, we refer to purple noise as the noise that has only high and low-frequency content, \ie, does not have a middle-frequency component. We apply a band-stop filter over the random noise $W$ to produce this type of noise. Specifically, we first apply a band-pass filter to $W$  to obtain green noise and then subtract it from the input $W$, preserving only the low and high frequencies. Formally, this transformation of the noise $W$ into \textit{\textcolor{mypurple}{purple noise}} $N_p$ can be expressed as:
\begin{equation}
N_p=W - (G_{\sigma_1}*W -G_{\sigma_2}*W),
\end{equation}
where $\sigma_1<\sigma_2$. Analyzing the periodogram in Fig. \ref{fig:proposed} (column ``Purple''), we can observe that this noise combines the characteristics of both red and blue noise.

\vspace{4pt}
\mysection{Mask Generation.} In implementation, we pre-compute \textit{color noise} offline and store them in GPU memory before initiating MAE pre-training. To efficiently generate the masks during pre-training, we first apply random transformations on the loaded noise tensor to get a $P$-sized square noise window for every image in the batch $B$, where $P$ is the total number of patches. Then, we select the highest values from the noise window according to the desired mask ratio. Specifically, we apply random crop, horizontal flip, and vertical flip image transformation. Note that these image transformations operate in the spatial domain; hence, the frequency properties described in the previous section are preserved \cite{gonzalez2008digital}. Algorithm \ref{algo:mask_gen} shows the pseudo-code for our masking approach in PyTorch style. 


\begin{algorithm}[t]
\scriptsize
\SetAlgoLined
    \PyCode{\textcolor{magenta}{import} torch}\\
    \BlankLine
    \PyCode{\textcolor{magenta}{def} mask\_generation(N,P,B,T,mask\_ratio):} \\
    \Indp   
    \PyComment{N: noise tensor ( \eg blue, green, purple or red noise).} \\
    \PyComment{P: total number of patches. B: batch size.} \\
    \PyComment{T: random crop, horizontal, and vertical flip PyTorch transforms.} \\
    \PyComment{mask\_ratio: the mask ratio of total patches (\eg 0.75).} \\
        \BlankLine
        \PyComment{apply random transforms (T) to get a $\sqrt{\text{P}}\times\sqrt{\text{P}}$ noise windows} \\
        \PyCode{windows = T(N)[:B]} \PyComment{Assuming B < N.shape[0]}\\
        \PyCode{len\_keep = int(P * (1 - mask\_ratio))}\\
        \PyCode{windows = windows.view(B, -1)}\\
        
        \BlankLine
        \PyComment{keep stronger values from the noise}\\
        \PyCode{ids\_shuffle = torch.argsort(windows, dim=1, descending=True)}\\
        \PyCode{ids\_restore = torch.argsort(ids\_shuffle, dim=1)}\\
        \PyCode{ids\_keep = ids\_shuffle[:, :len\_keep]}\\
        
        \BlankLine
        \PyComment{generate the binary mask: 0 is keep, 1 is remove}\\
        \PyCode{mask = torch.ones([B, P])}\\
        \PyCode{mask[:, :len\_keep] = 0}\\

        \BlankLine
        \PyComment{unshuffle to get the binary mask}\\
        \PyCode{mask = torch.gather(mask, dim=1, index=ids\_restore)}\\
        \PyCode{\textcolor{magenta}{return} mask, ids\_restore, ids\_keep}
\caption{Pseudo-Code of our masking approach in PyTorch style.}
\label{algo:mask_gen}
\end{algorithm}

Figure \ref{fig:proposed} shows examples of the generated masks for each color noise (see \protect\coloredcircled{darkgray}{\scriptsize M}). As observed, the produced masks have a particular pattern associated with the frequency properties of each noise. For example, blue noise produces binary masks whose values are distributed uniformly but without large empty areas or overly dense clusters. On the other hand, the masks produced by green noise can be seen as a ``clustered'' version of the masks produced with blue noise.

\section{Experiments}
\label{sec:experiments}

\mysection{Implementation Details.} We evaluate the performance of our proposed masking strategies under self-supervised pre-training with MAE \cite{he2022masked} on the ImageNet-1K \cite{russakovsky2015imagenet} dataset. Unless otherwise specified, we mainly use the standard ViT-B/16 \cite{dosovitskiy2020image} as the backbone, and the decoder consists of 8 Transformer layers with a hidden dimension 512. The input images are resized to $224 \times 224$, and the patch size is $16 \times 16$; thus, the resulting total sequence length is $L=196$. 
To make a fair comparison with the original MAE pre-trained with random masking \cite{he2022masked}, we use the same masking ratio of 75\%. 
Please refer to our supplementary material for additional details on implementation and experiments with different masking ratios. We evaluate transfer learning performance using our pre-trained \colormae~models on different datasets and downstream tasks described as follows:

\mysection{ImageNet Classification.} We evaluate the performance of MAE pre-trained with our proposed masking strategy on ImageNet-1K \cite{russakovsky2015imagenet} classification following the standard protocol \cite{he2022masked}. We perform end-to-end fine-tuning for 100 epochs and report the Top-1 accuracy (\%) obtained on the validation set. We maintain the same resolution of $224\times224$ on both pre-training and fine-tuning.

\mysection{COCO Object Detection and Instance Segmentation.} We employ ViTDet \cite{li2022exploring} as our object detector model, which utilizes a Vision Transformer backbone to perform object detection and instance segmentation. Unless otherwise specified, we perform end-to-end fine-tuning on the COCO dataset \cite{lin2014microsoft}, resizing the images to a resolution of $768\times768$ to expedite the fine-tuning process. We report the box average precision (AP$^{bbox}$) for object detection and the mask AP for instance segmentation (AP$^{mask}$).


\mysection{ADE20k Semantic Segmentation.} We employ UperNet \cite{xiao2018unified} as our segmentation model and perform end-to-end fine-tuning on the ADE20k \cite{zhou2017scene} dataset for 160k iterations with an image resolution of $512\times512$. The evaluation metric used is the mean Intersection over Union (mIoU)\cite{everingham2015pascal}.

\subsection{Exploring Masking Strategies Performance}
\label{sec:explore}

\begin{figure}[t]
  \centering
  \includegraphics[width=\columnwidth]{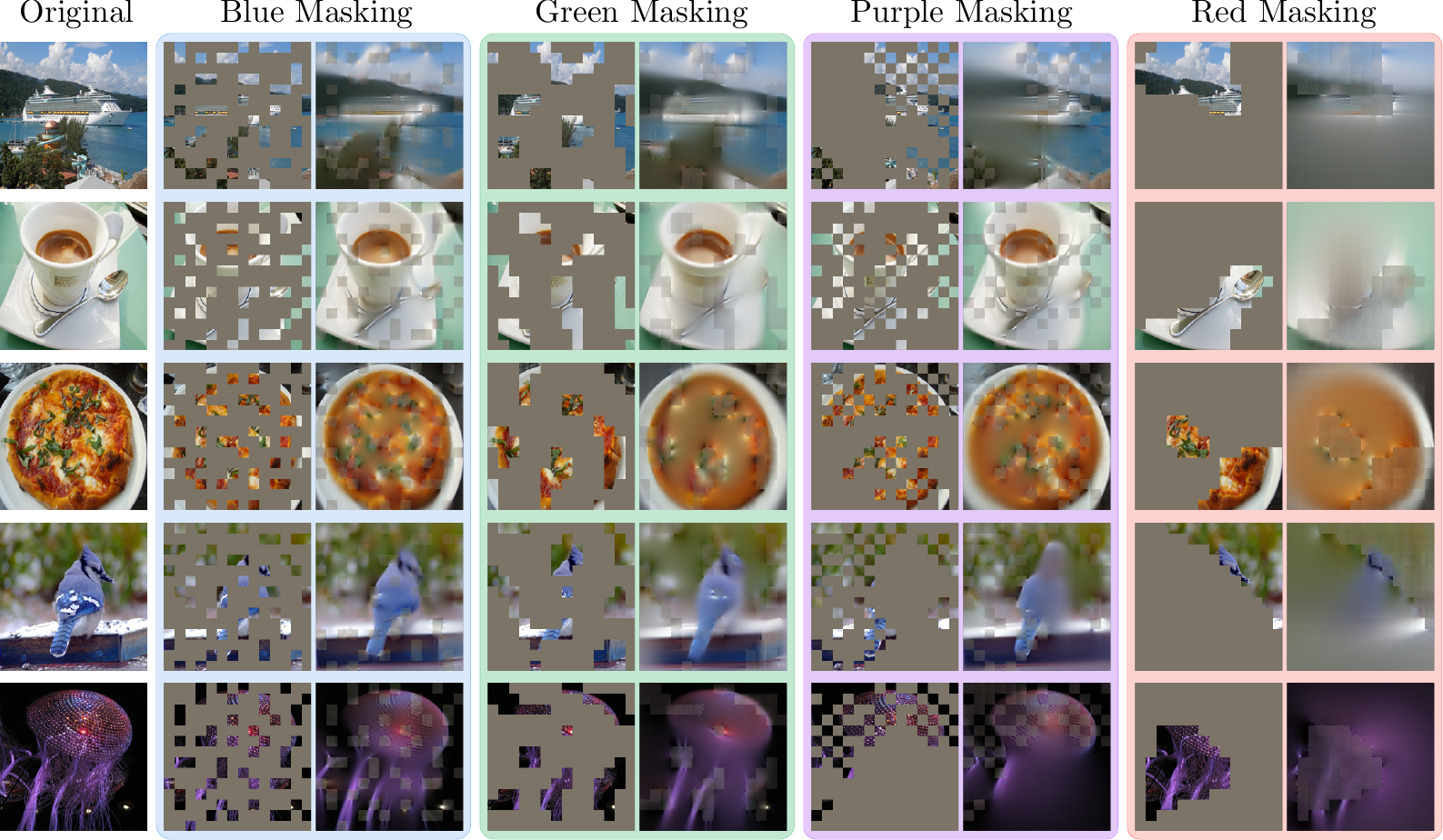}
  \caption{Reconstruction results on ImageNet validation images from MAE pre-trained during 300 epochs with our four generated masks: Blue, Green, Purple, and Red.}
  \label{fig:mask_vis}
\end{figure}

\mysection{Qualitative Results.} In this section, we evaluate the performance of MAE on the downstream tasks mentioned in the previous section when we pre-train with our proposed four types of \colormae ~masks. Figure \ref{fig:mask_vis} presents visualizations of the ImageNet validation images reconstructed 
with our four types of masks: Blue, Green, Purple, and Red. 
All the masks have the same masking ratio of 75\%, and the MAE model is pre-trained for 300 epochs to reconstruct unnormalized pixels. As observed, Blue masking provides better reconstruction quality as the reconstruction task is easy for the decoder. Similarly, while Green masking yields a lower reconstruction quality than blue, it still provides semantically meaningful and sufficiently accurate reconstructions. Conversely, the reconstruction task becomes significantly more challenging with Purple and Red masking, resulting in lower reconstruction quality. Specifically, Red masking is notably more ``aggressive'', leading to poor reconstruction and representation learning. Please refer to our supplementary for more reconstruction visualizations.

\begin{table}[t]
\caption{Downstream tasks performance after fine-tuning. MAE is pre-trained on ImageNet-1K \cite{russakovsky2015imagenet} with random masking and our proposed masking approach. We report ImageNet-1K Top-1 accuracy, ADE20K mIoU \cite{zhou2017scene}, and COCO AP$^{bbox}$ \cite{lin2014microsoft} for classification, semantic segmentation, and object detection, respectively.}
\resizebox{\linewidth}{!}{%
\begin{tabular}{c|ccccc|ccccc|ccccc}
\hline
                & \multicolumn{5}{c|}{Classification (Top-1 accuracy)}                                                                                                      & \multicolumn{5}{c|}{Semantic Segmentation (mIoU)}                                                                                                & \multicolumn{5}{c}{Object Detection (AP$^{bbox}$)}                                                                                                                    \\ \hline
Pretrain Epochs & Random & \cellcolor[HTML]{DAE8FC}Blue           & \cellcolor[HTML]{C4E7D4}Green          & \cellcolor[HTML]{E3CAFA}Purple & \cellcolor[HTML]{FED2CF}Red   & Random & \cellcolor[HTML]{DAE8FC}Blue  & \cellcolor[HTML]{C4E7D4}Green          & \cellcolor[HTML]{E3CAFA}Purple & \cellcolor[HTML]{FED2CF}Red   & Random         & \cellcolor[HTML]{DAE8FC}Blue           & \cellcolor[HTML]{C4E7D4}Green          & \cellcolor[HTML]{E3CAFA}Purple & \cellcolor[HTML]{FED2CF}Red   \\ \hline
100             & 81.69  & \cellcolor[HTML]{DAE8FC}\textbf{81.82} & \cellcolor[HTML]{C4E7D4}\textbf{81.82} & \cellcolor[HTML]{E3CAFA}80.82  & \cellcolor[HTML]{FED2CF}78.83 & 42.20  & \cellcolor[HTML]{DAE8FC}40.33 & \cellcolor[HTML]{C4E7D4}\textbf{42.24} & \cellcolor[HTML]{E3CAFA}38.22  & \cellcolor[HTML]{FED2CF}35.31 & 45.90          & \cellcolor[HTML]{DAE8FC}\textbf{46.00} & \cellcolor[HTML]{C4E7D4}45.90          & \cellcolor[HTML]{E3CAFA}44.10  & \cellcolor[HTML]{FED2CF}40.80 \\
300             & 82.82  & \cellcolor[HTML]{DAE8FC}82.56          & \cellcolor[HTML]{C4E7D4}\textbf{82.98} & \cellcolor[HTML]{E3CAFA}82.39  & \cellcolor[HTML]{FED2CF}81.35 & 44.51  & \cellcolor[HTML]{DAE8FC}43.42 & \cellcolor[HTML]{C4E7D4}\textbf{45.80} & \cellcolor[HTML]{E3CAFA}43.85  & \cellcolor[HTML]{FED2CF}42.08 & 48.50          & \cellcolor[HTML]{DAE8FC}48.10          & \cellcolor[HTML]{C4E7D4}\textbf{48.70} & \cellcolor[HTML]{E3CAFA}47.20  & \cellcolor[HTML]{FED2CF}45.10 \\
800             & 83.17  & \cellcolor[HTML]{DAE8FC}83.02          & \cellcolor[HTML]{C4E7D4}\textbf{83.57} & \cellcolor[HTML]{E3CAFA}82.92  & \cellcolor[HTML]{FED2CF}82.41 & 46.46  & \cellcolor[HTML]{DAE8FC}44.81 & \cellcolor[HTML]{C4E7D4}\textbf{49.18} & \cellcolor[HTML]{E3CAFA}45.96  & \cellcolor[HTML]{FED2CF}44.78 & 49.15 & \cellcolor[HTML]{DAE8FC}49.10          & \cellcolor[HTML]{C4E7D4}\textbf{49.50} & \cellcolor[HTML]{E3CAFA}48.50  & \cellcolor[HTML]{FED2CF}46.90 \\
1600            & 83.43  & \cellcolor[HTML]{DAE8FC}83.26          & \cellcolor[HTML]{C4E7D4}\textbf{83.77} & \cellcolor[HTML]{E3CAFA}83.20  & \cellcolor[HTML]{FED2CF}82.73 & 47.46  & \cellcolor[HTML]{DAE8FC}46.35 & \cellcolor[HTML]{C4E7D4}\textbf{49.26} & \cellcolor[HTML]{E3CAFA}47.23  & \cellcolor[HTML]{FED2CF}46.08 & 49.60          & \cellcolor[HTML]{DAE8FC}49.50          & \cellcolor[HTML]{C4E7D4}\textbf{50.10} & \cellcolor[HTML]{E3CAFA}49.10  & \cellcolor[HTML]{FED2CF}47.20 \\ \hline
\end{tabular}}
\label{tab:main_quantitative}
\end{table}

\mysection{Quantitative Results.} In Tab. \ref{tab:main_quantitative}, we investigate the performance of our four types of masks generated by our approach on various downstream tasks and show the comparison with traditional random masking. 
Our findings indicate that the Purple and Red masks exhibit lower performance, with the latter being the worst, unable to learn representations effectively. This aligns with the visualizations in Fig. \ref{fig:mask_vis}, where it clearly shows that Red masking significantly increases the difficulty of the reconstruction task. Note that although Blue masking provides the best (lower) reconstruction loss among all the approaches (Fig. \ref{fig:loss}), it does not yield better performance on downstream tasks (Tab. \ref{tab:main_quantitative}).
Finally, it is important to highlight that Green masking delivers the best results across all the evaluated downstream tasks. The improvement is particularly significant in the semantic segmentation task, with a notable increase of 2.72 in the mIoU metric compared to random masking when pre-training MAE for 800 epochs. We also observe that while Green masking consistently provides performance improvements, such enhancements become more evident when the pre-training epochs increase. Notably, there is a marked improvement in mIoU (3.38\textcolor{mygreen}{$\uparrow$}) when increasing the training from 300 to 800 epochs, which suggests our approach provides faster convergence than traditional random masking.

\begin{table}[t]
  \centering
  \caption{Additional MAE experiments with (a) ViT Large (ViT-L/16 \cite{dosovitskiy2020image}) as backbone and evaluated on two downstream tasks: ImageNet-1K classification and ADE20K semantic segmentation; (b) ViT Base (ViT-B/16 \cite{dosovitskiy2020image}) as backbone and evaluated on COCO object detection when using images with $768\times768$ and $1024\times1024$ resolution.}
  \label{tab:add_exps}
  \begin{subtable}{0.5\textwidth}
    \centering
    \resizebox{\columnwidth}{!}{
    \begin{tabular}{c|c|ccc|ccc}
    \hline
    & & \multicolumn{3}{c|}{ImageNet-1K (Top 1 Acc) } & \multicolumn{3}{c}{ADE20K (mIoU)} \\ \cline{3-8} 
    \multirow{-2}{*}{\begin{tabular}[c]{@{}c@{}}Pretrain \\ Epochs\end{tabular}} & \multirow{-2}{*}{Arch} & Random & \cellcolor[HTML]{DAE8FC}Blue & \cellcolor[HTML]{C4E7D4}Green & Random & \cellcolor[HTML]{DAE8FC}Blue & \cellcolor[HTML]{C4E7D4}Green \\ \hline
    300 & & 84.76 & \cellcolor[HTML]{DAE8FC}84.77 & \cellcolor[HTML]{C4E7D4}\textbf{85.02} & 47.55 & \cellcolor[HTML]{DAE8FC}46.75 & \cellcolor[HTML]{C4E7D4}\textbf{49.00} \\
    800 & \multirow{-2}{*}{\begin{tabular}[c]{@{}c@{}}ViT \\ Large\end{tabular}} & 85.42 & \cellcolor[HTML]{DAE8FC}85.34 & \cellcolor[HTML]{C4E7D4}\textbf{85.64} & 50.29 & 49.38 \cellcolor[HTML]{DAE8FC} & \cellcolor[HTML]{C4E7D4}\textbf{51.46} \\ \hline
    \end{tabular}}
    \caption{Downstream tasks performance with MAE pre-trained using ViT-Large as the backbone and using random, Blue, and Green masking.}
  \end{subtable}
  \hfill
  \begin{subtable}{0.45\textwidth}
    \centering
    \resizebox{\columnwidth}{!}{
    \begin{tabular}{c|c|ccc|ccc}
    \hline
    & & \multicolumn{3}{c|}{Image Size 768x768} & \multicolumn{3}{c}{Image Size 1024x1024} \\ \cline{3-8} 
    \multirow{-2}{*}{\begin{tabular}[c]{@{}c@{}}Pretrain \\ Epochs\end{tabular}} & \multirow{-2}{*}{Arch} & Random & \cellcolor[HTML]{DAE8FC}Blue & \cellcolor[HTML]{C4E7D4}Green & Random & \cellcolor[HTML]{DAE8FC}Blue & \cellcolor[HTML]{C4E7D4}Green \\ \hline
    300 & & 48.50 & \cellcolor[HTML]{DAE8FC}48.10 & \cellcolor[HTML]{C4E7D4}\textbf{48.70} & 50.10 & \cellcolor[HTML]{DAE8FC}49.80 & \cellcolor[HTML]{C4E7D4}\textbf{50.40} \\
    1600 & \multirow{-2}{*}{\begin{tabular}[c]{@{}c@{}}ViT \\ Base\end{tabular}} & 49.60 & \cellcolor[HTML]{DAE8FC}49.50 & \cellcolor[HTML]{C4E7D4}\textbf{50.10} & 50.90 & \cellcolor[HTML]{DAE8FC}50.80 & \cellcolor[HTML]{C4E7D4}\textbf{51.50} \\ \hline
    \end{tabular}}
    \caption{Object detection performance when fine-tuning using images resized to $768\times 768$ \vs images with $1024 \times 1024$ resolution.}
  \end{subtable}
  \vspace{-20pt}
\end{table}

\mysection{Additional Results.} Tab. \ref{tab:add_exps} (a) presents additional experiments when pre-training MAE using ViT Large (ViT-L/16 \cite{dosovitskiy2020image}) as a backbone and evaluate downstream performance on ImageNet-1K classification task and semantic segmentation on ADE20K dataset. The table shows that the results are consistent with those presented in Tab. \ref{tab:main_quantitative}. Specifically, Green masking performs best in both tasks, outperforming random and Blue masking. Additionally, Tab. \ref{tab:add_exps} (b) presents a comparison between object detection performance when fine-tuning on the COCO dataset using images resized to $768\times768$ resolution versus $1024 \times 1024$ resolution. This experiment uses ViT Base (ViT-B/16 \cite{dosovitskiy2020image}) as the backbone for MAE. The performance enhancements obtained with Green masking are consistent with previous results, outperforming random and Blue masking.

\subsection{Comparison with Other Methods}

\begin{table}[t]
\caption{Comparison with state-of-the-art methods pre-trained on ImageNet-1K. The resolution of images is $224\times224$ for both pre-training and fine-tuning. $\dagger$ indicates our implementation, including pre-training and fine-tuning. $\ddagger$ means the results are borrowed from \cite{chen2024context}. $\mathsection$ means the results are borrowed from \cite{wang2024droppos}.}
\resizebox{\columnwidth}{!}{
\begin{tabular}{lcc|c|c|cccccc}
\toprule
                          &                                                                            &                                                                              & ADE20K        & ImageNet   & \multicolumn{6}{c}{COCO}                                                                                                               \\ \cline{4-11} 
\multirow{-2}{*}{Method}  & \multirow{-2}{*}{\begin{tabular}[c]{@{}c@{}}Pretrain\\ Epoch\end{tabular}} & \multirow{-2}{*}{\begin{tabular}[c]{@{}c@{}}Pre-trained\\ Data\end{tabular}} & mIoU          & Top-1 Acc. & AP$^{bbox}$ & AP$^{bbox}_{50}$ & \multicolumn{1}{c|}{AP$^{bbox}_{75}$}             & AP$^{mask}$ & AP$^{mask}_{50}$ & AP$^{mask}_{75}$ \\ \midrule
\multicolumn{3}{l}{\textit{Non-MIM}} \\
MoCo v3 $^{\ddagger}$ \cite{Chen_2021_ICCV}     & 600                                                                        & IN1K                                                                         & 47.2          & 83.0       & 45.5        & 67.1             & \multicolumn{1}{c|}{49.4}                         & 40.5        & 63.7             & 43.4             \\
DINO $^{\ddagger}$ \cite{caron2021emerging}        & 1600                                                                       & IN1K                                                                         & 47.2          & 83.3       & 46.8        & 68.6             & \multicolumn{1}{c|}{50.9}                         & 41.5        & 65.3             & 44.5             \\
DropPos\cite{wang2024droppos}                   & 800                                                                        & IN1K                                                                         & 47.8          & 84.2       & 47.7        & 68.3             & \multicolumn{1}{c|}{52.8}                         & 42.6        & 65.3             & 46.2             \\ 
\midrule
\multicolumn{3}{l}{\textit{MIM with data-adaptive masking}}   \\
AttMask  \cite{kakogeorgiou2022hide} & 100                                                                        & IN1K                                                                         & 45.3          & -          & 48.8        & -                & \multicolumn{1}{c|}{-}                            & 42.0        & -                & -                \\
UM-MAE \cite{li2022uniform}  & 200                                                                        & IN1K                                                                         & 42.6         & 82.9       & 45.9        & 64.5             & \multicolumn{1}{c|}{50.2}                         & -           & -                & -                \\
SemMAE$^{\mathsection}$  \cite{li2022semmae} & 800                                                                        & IN1K                                                                         & 44.9          & 83.4       & 45.6        & 66.2             & \multicolumn{1}{c|}{55.2}                         & 40.9        & 63.3             & 44.4             \\
HPM   \cite{wang2023hard}  & 800                                                                        & IN1K                                                                         & 48.5          & 84.2       & 50.1        & -                & \multicolumn{1}{c|}{-}                            & 44.6        & -                & -                \\
\midrule

\multicolumn{3}{l}{\textit{MIM with data-independent masking}}  \\
BEiT~\cite{bao2021beit}                      & 800                                                                        & IN1K+DALLE                                                                          & 45.6          & 83.2       & 40.8        & 59.4             & \multicolumn{1}{c|}{44.1}                         & 36.0        & 56.8             & 38.2             \\
MAE$^{\dagger}$ \cite{he2022masked} & 800                                                                        & IN1K                                                                         & 46.5          & 83.2       & 49.2        & 69.7             & \multicolumn{1}{c|}{53.9}                         & 43.4        & 66.6             & 46.9             \\
MixedAE  \cite{chen2023mixed} & 800                                                                        & IN1K                                                                         & 48.7          & 83.5       & 50.3        & 69.1             & \multicolumn{1}{c|}{54.8}                         & 43.5        & 66.2             & 47.4             \\
\rowcolor[HTML]{C4E7D4}ColorMAE-G             & 800                                                                        & IN1K                                                                         & 49.2          & 83.6       & 49.5        & 70.0             & \multicolumn{1}{c|}{\cellcolor[HTML]{C4E7D4}54.2} & 43.7        & 67.1             & 47.1             \\
MAE \cite{he2022masked} & 1600   & IN1K       & 48.1          &83.6       & \textbf{50.6}        & 69.4             & \multicolumn{1}{c|}{\textbf{55.0}}   &43.8        &66.6             & 47.5             \\

\rowcolor[HTML]{C4E7D4}ColorMAE-G              & 1600                                                                       & IN1K                                                                         & \textbf{49.3} & \textbf{83.8}       & 50.1       & \textbf{70.7}             & \multicolumn{1}{c|}{\cellcolor[HTML]{C4E7D4}54.7} & \textbf{44.4}        & \textbf{67.8}             & \textbf{48.0}             \\ 
\bottomrule
\end{tabular}}
\label{table:sota_cmp}
\end{table}

Given the results from the previous section, we will focus this section on comparing Green masks with other methods. Here, we refer to MAE pre-trained with Green masks as \textcolor{mygreen}{ColorMAE-G}. 

\mysection{Comparison with MAE.} The bottom block of Tab. \ref{table:sota_cmp} showcases that our \textcolor{mygreen}{ColorMAE-G} consistently outperforms MAE$^{\dagger}$ (our implementation), as well as the results presented in the original MAE paper \cite{he2022masked}, without incurring additional overhead (see Tab. \ref{tab:complexity}). The most notable enhancement is observed in the semantic segmentation task, where there is a significant improvement of $+2.7$ mIoU with 800 pretraining epochs. Furthermore, our method also demonstrates competitive or superior results in object detection tasks.

\mysection{Comparison with Other Data-Independent Masking Methods.} In the last block of Tab. \ref{table:sota_cmp}, we also compare our approach against other state-of-the-art data-independent masking methods, which usually use random or block-wise masking. As observed, \textcolor{mygreen}{ColorMAE-G} outperforms all these methods in the semantic segmentation task on the ADE20K dataset and provides comparable results in the other downstream tasks, where it is only surpassed by MixedAE ($-0.2$) and MAE ($-0.5$) in COCO object detection.

\mysection{Comparison with Data-adaptive Masking Methods.} Notably, our data-independent masking approach also achieves competitive performance even compared to sophisticated data-adaptive masking, which incorporates additional attention-based or adversarial-guided mechanisms, increasing the computational cost. In the semantic segmentation task on the ADE20K dataset, our approach visibly outperforms these methods. Our performance in the object detection task is also better than these approaches and comparable to HPM~\cite{wang2023hard}. However, we do not introduce any additional parameters or computations in the network, thus enjoying the benefit of fast training (see Tab. \ref{tab:complexity}).





\subsection{Analysis}
\label{sec:analysis}

\begin{figure}[t]
\centering
\begin{minipage}{0.45\textwidth}
  \centering
  \includegraphics[width=\linewidth]{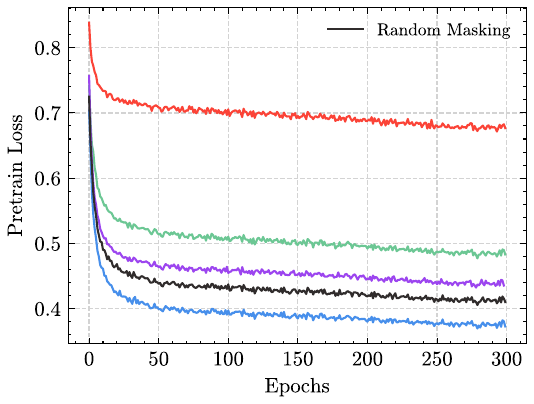} 
  \caption{MAE pre-training loss for different masking strategies with ViT-B.}
  \label{fig:loss}
\end{minipage}\hfill
\begin{minipage}{0.5\textwidth}
  \centering
  \vspace{-0.5in}
  \captionof{table}{Complexity analysis of the MAE model when pre-training with traditional random masking and our proposed masking strategies.}
  \vspace{5pt}
  \resizebox{\columnwidth}{!}{
  \begin{tabular}{@{}ccccc@{}}
    \toprule
    \begin{tabular}[c]{@{}c@{}}Masking \\ Strategy\end{tabular} & \begin{tabular}[c]{@{}c@{}}Parameters \\ (M)\end{tabular} & \begin{tabular}[c]{@{}c@{}}Flops \\ (G)\end{tabular} & \begin{tabular}[c]{@{}c@{}}Memory \\ (GB)\end{tabular} & \begin{tabular}[c]{@{}c@{}}Pre-training Time \\ per Epoch (Min)\end{tabular} \\ \midrule
    Random                                                      & 111.91                                                    & 16.87                                                & 27.44                                                  & 5.21                                                                         \\
    \rowcolor[HTML]{DAE8FC} 
    Blue                                                        & 111.91                                                    & 16.87                                                & 28.21                                                  & 5.18                                                                         \\
    \rowcolor[HTML]{C4E7D4} 
    Green                                                       & 111.91                                                    & 16.87                                                & 28.21                                                  & 5.18                                                                         \\
    \rowcolor[HTML]{E3CAFA} 
    Purple                                                      & 111.91                                                    & 16.87                                                & 28.21                                                  & 5.18                                                                         \\
    \rowcolor[HTML]{FED2CF} 
    Red                                                         & 111.91                                                    & 16.87                                                & 28.21                                                  & 5.18                                                                         \\ \bottomrule
    \end{tabular}}
  \label{tab:complexity}
\end{minipage}
\end{figure}

\mysection{Reconstruction Loss \vs Downstream Performance.} Figure \ref{fig:loss} presents the MAE pre-training (reconstruction) loss curves for random masking and our proposed masking approach. As observed, Blue masking achieves the lowest reconstruction loss over the epochs, followed by random, Purple, Green, and Red masking. Interestingly, Green masking does not yield the lowest pre-training loss, yet it achieves the best performance in downstream tasks, see Tab. \ref{tab:main_quantitative}. Our results contradict the hypothesis that a lower reconstruction loss implies better downstream performance \cite{zhang2022mask}. However, our findings resonate with the observations in \cite{wang2023hard}. Specifically, considering the reconstruction loss as a metric of the difficulty of the pre-training task, authors in \cite{wang2023hard} propose to mask the patches with higher loss, increasing downstream performance. Such hard-to-reconstruct patches are usually associated with the discriminative parts of an image, like objects. Authors in \cite{wang2023hard} conclude that consistently increasing the difficulty of the pretext task does not lead to better performance, and retaining a certain degree of \textit{randomness} is necessary for better results. Similarly, as observed from Fig. \ref{fig:loss}, our Red masking approach tends to mask out big segments of the image, making the pretext task very difficult but not allowing the model to learn useful feature representations. On the other hand, our Green masking approach masks out smaller random segments in the image, making the pretext task difficult enough to learn better representations. In general, our Green masking provides a better balance between pre-training task difficulty and randomness.













\mysection{Computational Cost.} In general, data-adaptive masking approaches inevitably increase the computational cost and number of parameters since they need to introduce additional components to the network, \eg, an extra decoder. For instance, HPM \cite{wang2023hard} increases the training time $1.1\times$, while CAE \cite{chen2024context} increases the number of parameters to $1.23\times$ and training time to $1.24\times$ in comparison with MAE \cite{he2022masked}. Similarly, authors in \cite{feng2023evolved} report that their mask generation occupies 12\% of pre-training time. On the other hand, our proposed data-independent masking strategy is efficient and does not add extra model parameters or computational overhead, as shown in Tab. \ref{tab:complexity}. Because we pre-compute the noise color patterns offline and store them in GPU memory, there is only a small increment in memory usage compared to the original MAE model. In particular, during our experiments, we use $3072$ noise patterns of $256\times256$ spatial dimension for each type of color noise, leading to a small increment of $2.8\%$ of memory. However, fewer patterns can be used, reducing memory costs while slightly impacting performance. Please see our supplementary for additional experiments when varying the number of noise patterns used during \colormae~pre-training.


\begin{figure}[t]
    \centering
    \includegraphics[width=\columnwidth]{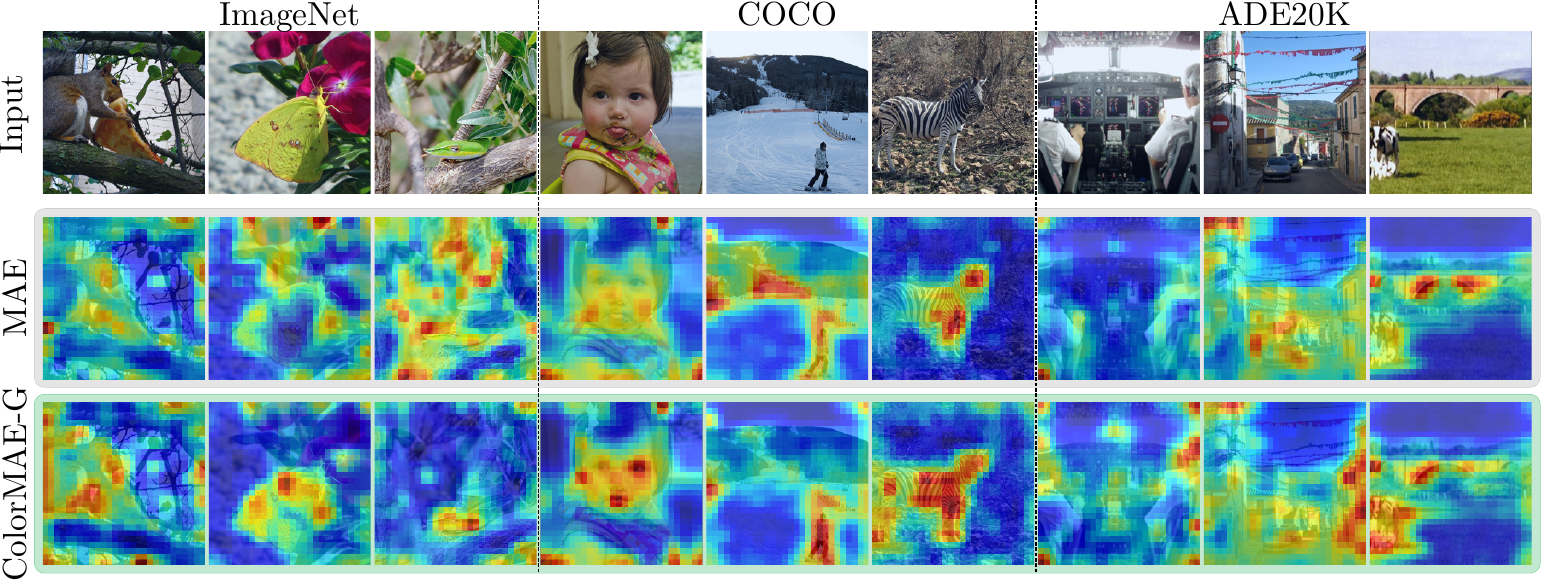}
    \caption{Self-attention of the [CLS] tokens averaged across the heads of the last layer in MAE pre-trained using random masking and our proposed Green masking approach (\textcolor{mygreen}{ColorMAE-G}). We show attention maps on images from Imagenet-1K\cite{russakovsky2015imagenet}(1st-3rd columns), Microsoft COCO \cite{lin2014microsoft}(4th-6th columns) and ADE20K \cite{zhou2017scene}(7th-9th columns) datasets. Both MAE and \textcolor{mygreen}{ColorMAE-G} are pre-trained on ImageNet-1K for 300 epochs. Please refer to our supplementary for more visualizations of the attention maps when pre-training MAE with other \colormae~masks.}
    \label{fig:attn_results}
\end{figure}

\mysection{Attention Analysis.} We show examples of self-attention maps of the [CLS] tokens averaged across the heads of the last layer in Fig. \ref{fig:attn_results} for the three different datasets. Here, we show the results for MAE pre-trained using random and our proposed Green masking approach. From the visualizations, our \greencolormae effectively identifies the foreground object patches with better precision and completeness.
This might also explain its superior performance when transferred to dense perception tasks such as semantic segmentation \cite{zhou2017scene}, object detection, and instance segmentation \cite{lin2014microsoft}.

\begin{figure}[t]
    \centering
    \includegraphics[width=\columnwidth]{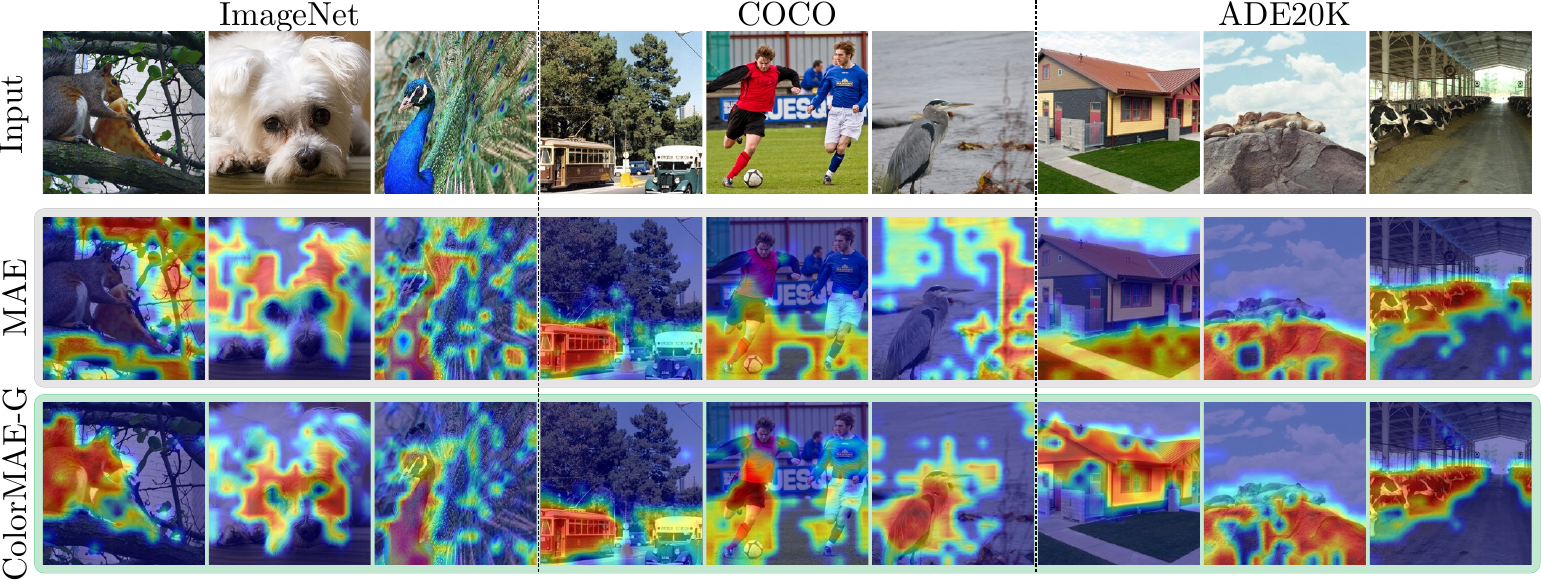}
    \caption{Comparative visualization of Class Activation Maps (CAM) generated with EigenCAM\cite{muhammad2020eigen} for ViT-B. First, we perform self-supervised pre-training using standard MAE with random masking and our \textcolor{mygreen}{ColorMAE-G} on the ImageNet-1K dataset. Then, we conduct end-to-end supervised fine-tuning following the standard protocol \cite{he2022masked} for ImageNet classification during 100 epochs. We show CAM maps of the ViT-B pre-trained with MAE (second row) and our \textcolor{mygreen}{ColorMAE-G} (third row) on images from ImageNet-1K (1st-3rd column), Microsoft COCO (4th-6th columns), and ADE20K (7th-9th columns) datasets. Please refer to our supplementary for more visualizations.}
    \label{fig:cam}
\end{figure}

Additionally, in Fig. \ref{fig:cam}, we employed EigenCAM \cite{muhammad2020eigen} to visualize the activation patterns of ViT-B, highlighting the model focus areas during image classification tasks. We first perform self-supervised pre-training on the ImageNet-1K dataset using MAE with random masking and our \greencolormae approach. Then, we conduct end-to-end supervised fine-tuning for 100 epochs
on the ImageNet classification task. The resulting CAMs, depicted in the second and third rows of Fig. \ref{fig:cam}, offer a visual comparison of the attention mechanisms of the models. As observed, the CAMs of ViT-B pre-trained with our \greencolormae exhibit more localized and relevant attention areas (\eg, discriminative objects/subjects in the scene), especially in contrast to those provided by ViT pre-trained with the standard MAE. 
Please refer to our supplementary document for more self-attention 
and CAM maps.

\section{Conclusions}
\label{sec:conclusions}

Until now, random masking has been the foundational strategy and a common starting point for developing data-adaptive masking strategies. This paper explored four distinct data-independent masking alternatives to the conventional random masking approach. Using our \colormae~approach, we can generate different random masks with specific patterns by using noise with different frequency spectra. We observed that by using our generated \textit{Green masks} during MAE pre-training (\textcolor{mygreen}{ColorMAE-G}) we achieved faster convergence and better performance in downstream tasks, especially in semantic segmentation. Among the explored data-independent approaches in this paper, we found that our Green masks provide the best balance between pretext task difficulty and randomness, which allows the model to learn better representations.

\mysection{Discussion and Limitations.} In this work, we adopted a simple yet effective approach to filter random noise and generate different masks. While we investigated additional algorithms, such as the void-and-cluster method \cite{ulichney1993void}, for generating improved blue noise patterns, these did not yield significant performance enhancements and resulted in slower mask generation. Similarly, developing or using other algorithms to produce better \textit{green noise} patterns \cite{lau1998green} can be explored in future works. On the other hand, while our method is computationally efficient, it increments memory usage by 2.8\%. Although this increment is small, it could be considered a limitation and could be improved in future works.

\mysection{Acknowledgments.} This work was supported by the KAUST Center of Excellence on GenAI under award number 5940.

\bibliographystyle{splncs04}
\bibliography{main}
\end{document}